\pdfoutput=1

\documentclass[11pt]{article}

\usepackage[preprint]{acl}

\usepackage{times}
\usepackage{latexsym}

\usepackage[T1]{fontenc}

\usepackage[utf8]{inputenc}

\usepackage{microtype}

\usepackage{inconsolata}

\usepackage{graphicx}
\usepackage{algorithm}
\usepackage{algorithmic}
\usepackage{makecell}
\usepackage{multirow}
\usepackage{longtable}
\usepackage{booktabs}
\usepackage{soul}
\usepackage{color, colortbl}
\definecolor{Gray}{gray}{0.9}
\usepackage{xcolor} 
\usepackage{amsmath}
\usepackage{todonotes}
\usepackage{changes}
\usepackage{enumitem}
\usepackage{amssymb,mathrsfs}
\usepackage{fancybox}
\usepackage{tcolorbox}
\newcommand{\up}[1]{\textcolor{red}{{$\uparrow$ #1}}}
\newcommand{\down}[1]{\textcolor{red}{{$\downarrow$ #1}}}
\definecolor{tabcolor5}{RGB}{210, 220, 250} 
\definecolor{tabcolor3}{RGB}{204, 232, 207} 
\newcommand{\aaa}{\mathbf{a}}
\newcommand{\xx}{\mathbf{x}}
\newcommand{\vv}{\mathbf{v}}
\newcommand{\R}{\mathbb{R}}
\newcommand{\paren}[1]{\left(#1\right)}
\usepackage{placeins}
\title{Contrastive Prompting Enhances Sentence Embeddings in LLMs\\ through Inference-Time Steering}

\author{%
\textbf{Zifeng Cheng}\footnotemark[1] \quad \textbf{Zhonghui Wang}\footnotemark[1] \quad \textbf{Yuchen Fu}\footnotemark[1] \quad \textbf{Zhiwei Jiang}\footnotemark[2] \\
\textbf{Yafeng Yin} \quad  \textbf{Cong Wang} \quad \textbf{Qing Gu}\\
State Key Laboratory for Novel Software Technology, Nanjing University, China \\
\texttt{\{chengzf,yuchenfu,zhonghuiwang\}@smail.nju.edu.cn}, \\
\texttt{\{jzw,yafeng\}@nju.edu.cn}, \texttt{cw@smail.nju.edu.cn, guq@nju.edu.cn} 
}

\renewcommand{\thefootnote}{\fnsymbol{footnote}}

\begin{document}

\maketitle
\begin{abstract}
Extracting sentence embeddings from large language models (LLMs) is a practical direction, as it requires neither additional data nor fine-tuning.
Previous studies usually focus on prompt engineering to guide LLMs to encode the core semantic information of the sentence into the embedding of the last token.
However, the last token in these methods still encodes an excess of non-essential information, such as stop words, limiting its encoding capacity.
To this end, we propose a Contrastive Prompting (CP) method that introduces an extra auxiliary prompt to elicit better sentence embedding. 
By contrasting with the auxiliary prompt, CP can steer existing prompts to encode the core semantics of the sentence, rather than non-essential information.
CP is a plug-and-play inference-time intervention method that can be combined with various prompt-based methods.
Extensive experiments on Semantic Textual Similarity (STS) tasks and downstream classification tasks demonstrate that our method can improve the performance of existing prompt-based methods across different LLMs.
Our code will be released at \url{https://github.com/zifengcheng/CP}.
\end{abstract}

\footnotetext[1]{Equal Contribution.}
\footnotetext[2]{Corresponding Author.}
\renewcommand{\thefootnote}{\arabic{footnote}}

\section{Introduction}
Sentence embeddings~\cite{DBLP:journals/corr/abs-2412-09165} play a fundamental role in various real-world applications, such as information retrieval, text classification, clustering, and so on.
Considering that large language models (LLMs) have achieved success in zero-shot settings across various tasks, some works~\cite{liumeaning,lei2024meta,tp} have started to directly extract sentence embeddings from the hidden states of LLMs without the need for additional data or fine-tuning.
Since data are likely to be scarce in practice and the cost of fine-tuning LLM is expensive, such zero-shot setting is more practical and promising, while preserving the general capabilities of LLMs.

\begin{figure}[t]
    \centering
    \includegraphics[width=0.48\textwidth,trim=4.1cm 1.07cm 14.5cm 5.2cm,clip]{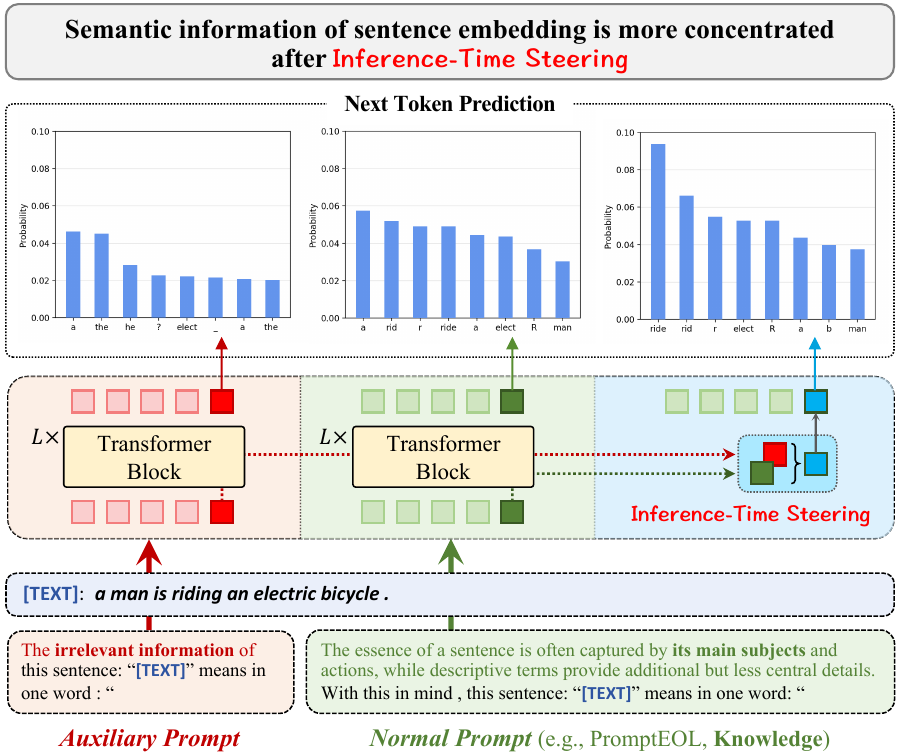}
    \caption{
    The comparison of sentence embeddings obtained by the auxiliary prompt, normal prompt, and our proposed inference-time steering method. 
    The decoding probabilities of Next Token Prediction are used to reflect the semantic information contained in the corresponding sentence embeddings.
    }
    \label{fig:moti}
\end{figure}

Existing works~\cite{jiang2023scaling,lei2024meta,zhang2024simple} typically focus on prompt engineering to compress the full semantics of a sentence into the last token and use the hidden state of that token as the sentence embedding.
PromptEOL~\cite{jiang2023scaling} first utilizes a simple and effective prompt: \textit{This sentence:} ``\texttt{[TEXT]}'' \textit{means in one word: ``}, to generate sentence embeddings, where \texttt{[TEXT]} serves as the sentence slot.
Subsequently, some works~\cite{lei2024meta,zhang2024simple} have designed various prompts to enable the last token to capture the core semantics of the sentence, rather than focusing on non-essential information.
Specifically, MetaEOL~\cite{lei2024meta} uses a diverse set of meta-task prompts.
Pretended CoT~\cite{zhang2024simple} employs a Chain-of-Thought prompt.
Knowledge~\cite{zhang2024simple} directs the LLM to focus more on the sentence's main subjects and actions.

Existing methods can be viewed as an \emph{indirect} approach, where different prefix prompts \emph{indirect} alter the representation of the last token, making it more focused on the core semantics of the sentence.
However, through the pilot next token prediction decoding experiment, their method still overly encodes non-essential information, such as stopwords, as shown in Figure~\ref{fig:moti}.
Although the Knowledge~\cite{zhang2024simple} prompt emphasizes grasping the main subjects and actions, the token with the highest probability remains the non-essential stopword ``a''.

In this paper, we propose a simple and effective \textbf{C}ontrastive \textbf{P}rompting (\textbf{CP}) method that can be combined with existing methods.
CP additionally introduces an auxiliary prompt to encode non-essential information of the sentence and to contrast it with the normal prompt (i.e., existing prompt-based methods), enabling CP to \emph{directly} modify the hidden state of the last token in the normal prompt during inference.
By contrasting with the auxiliary prompt, CP can steer existing prompts to encode the core semantics of the sentence while filtering out non-essential information.
Specifically, we first forward propagate the sentences wrapped with the auxiliary and normal prompts to the specific layer and extract the sentence embeddings.
Then, we directly compare the embeddings of the two sentences and replace the representation of the last token in the normal prompt to steer its focus toward the core semantics of the sentence.
Due to the change in the norm of the sentence embedding before and after the intervention, we propose two strategies to control the norm of the sentence embedding after the intervention.
Finally, we continue to forward propagate the normal prompt to obtain the sentence embeddings.

Our main contributions are as follows:
\begin{itemize}
\item We first propose enhancing the quality of sentence embeddings through inference-time activation steering.
\item We propose the Contrastive Prompting method, which guides LLMs to encode the core semantics of a sentence into its embedding. Several specifically designed auxiliary prompts are explored, and two norm adjustment strategies are introduced during activation steering.
\item We conduct extensive experiments on Semantic Textual Similarity (STS) benchmarks and downstream classification tasks. Experimental results demonstrate that our proposed method significantly improves the performance of existing prompt-based methods across different LLMs. Additionally, since the auxiliary prompt only needs to propagate to the lower layers of LLMs, the extra time overhead is relatively minimal.
\end{itemize}

\section{Related Work}
\textbf{Sentence Embeddings} Sentence embedding aims to represent the semantic information of a sentence into a fixed-size vector representation and plays a fundamental role in various applications~\cite{MPD,DENN,DBLP:conf/iclr/0034TGSJ0Z25}.
Previous methods often focus on various data augmentation techniques and contrastive losses to fine-tune smaller pre-trained language models for enhancing sentence embeddings~\cite{gao2021simcse,DBLP:conf/emnlp/JiangJHZWZWHDZ22,DBLP:conf/acl/NiACMHCY22,DBLP:conf/acl/ChanchaniH23,DBLP:conf/acl/SuSKWHOYSZ023,shen2024imagpose,shen2025imagdressing}.
Due to the exceptional capabilities of LLMs, recent works~\cite{li2024bellm,behnamghader2024llm2vec,NV-Embed/Lee,generative/muennighoff} have begun fine-tuning LLMs to obtain sentence embeddings.
In addition, some studies~\cite{ESE,starbucks} focus on extracting scalable sentence embeddings from the intermediate layers of language models.
However, these methods require data and fine-tuning, leading to a high cost and a loss of LLMs' other general capabilities.
Thus, this paper focuses on directly extracting sentence embeddings from LLMs without the need for fine-tuning or data.

\textbf{Extracting Sentence Embeddings from LLMs}
Existing methods on extracting sentence embeddings from LLMs mainly focus on prompt engineering.
PromptEOL \cite{jiang2023scaling} first demonstrates the potential of LLMs in generating sentence embeddings by leveraging prompt engineering and compressing the semantics of a sentence into a single token.
Echo embeddings \cite{springer2024repetition} repeat the input twice within the context, allowing later tokens to access earlier tokens, and extract embeddings from the second occurrence.
MetaEOL \cite{lei2024meta} designs meta-task prompts via ChatGPT-4 to guide LLMs to consider sentence representations from multiple perspectives.
Pretended CoT \cite{zhang2024simple} uses CoT to inspire LLMs to output better embeddings.
Knowledge Enhancement \cite{zhang2024simple} directs the LLM to focus more on the sentence's main subjects and actions through prompts.
In this paper, we propose a plug-and-play method to further improve the various prompt-based methods.

\begin{figure*}[t]
    \centering
    \includegraphics[width=1\textwidth,trim=3.1cm 2.9cm 3.9cm 5.7cm,clip]{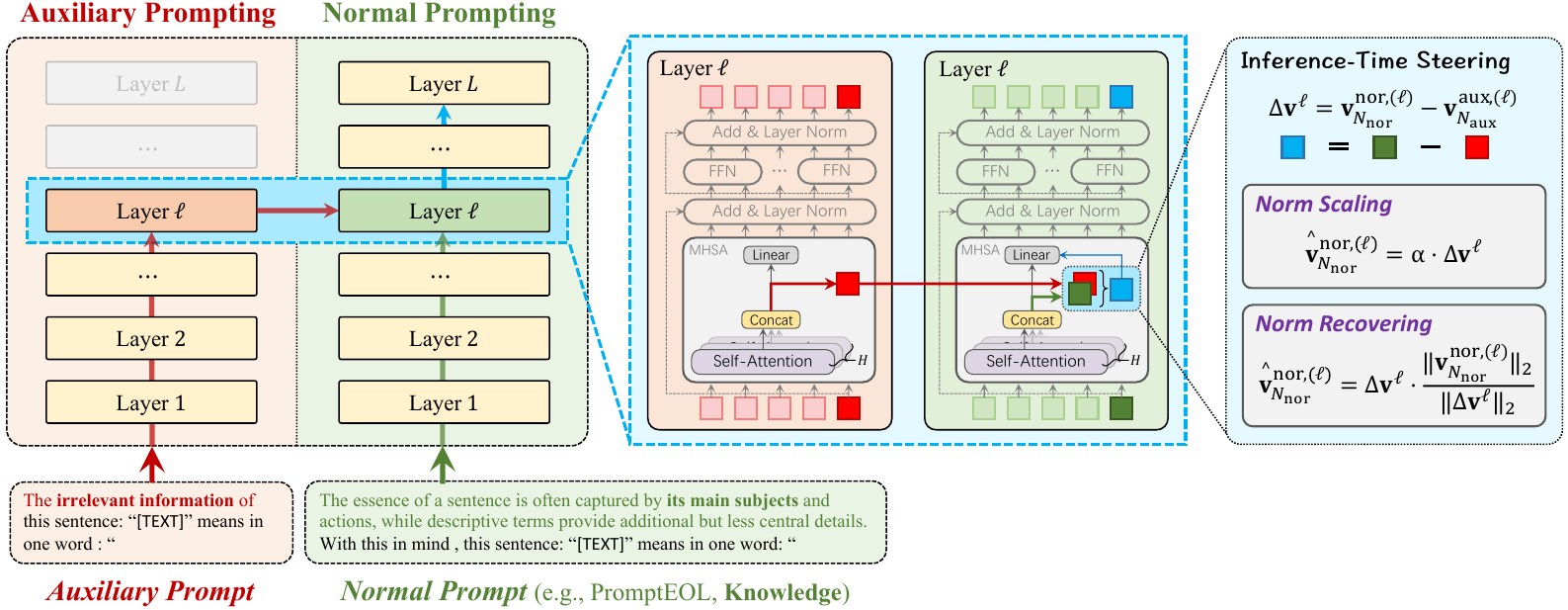}
    \caption{Illustration of the contrastive prompting method.}
    \label{fig:framework}
\end{figure*}

\textbf{Activation Steering}
Activation steering~\cite{RE,DBLP:conf/acl/RimskyGSTHT24,DBLP:conf/nips/0002PVPW23,DBLP:conf/emnlp/LeongCWWL23} creates a steering vector to modify the activations of the LLM, thereby controlling its generation.
The steering vector is derived by calculating the difference between activations from pairs of positive and negative supervision samples.
In contrast to these methods, we generate an activation vector for each sentence using two different prompts to refine the representation of the normal prompt, without the need for supervised data.

\section{Preliminary}
\paragraph{Extracting Embeddings from LLMs}
Previous work mainly focuses on prompt engineering to extract sentence embeddings from LLMs. 
PromptEOL~\cite{jiang2023scaling} introduces a widely adopted template for extracting sentence embeddings from LLMs:
\begin{equation}
    \text{This sentence: ``\texttt{[TEXT]}'' means in one word: ``} \nonumber
\end{equation}
where \texttt{[TEXT]} denotes the placeholder for the input sentence and the hidden state of the last token `` is considered as the sentence embedding.
The phrase ``in one word'' is a constraint that can prevent LLMs from generating long sentences, limiting a sentence to being represented by the embedding of a single word.

\paragraph{Multi-Head Attention in LLMs}
The $\ell$-th multi-head attention layer~\cite{transformers} contains three projection matrices $W^\ell_Q,W^\ell_K,W^\ell_V \in \R^{d \times d}$ and an output matrix $W^\ell_O \in \R^{d \times d}$.
The columns of each projection matrix and the rows of the output matrix can be split into $H$ heads, yielding $W^{\ell,h}_Q,W^{\ell,h}_K,W^{\ell,h}_V \in \R^{d \times \frac{d}{H}}$ for $h \in \left[1, H\right]$.
The $h$-th attention head computes the attention weight matrix $A^{\ell,h}\in \R^{N\times N}$ as follows:
\begin{align}
    \resizebox{\linewidth}{!}{$
    A^{\ell,h} = \varphi\paren{
    \frac{
        \paren{\xx^{\ell-1} W^{\ell,h}_Q}
        \paren{\xx^{\ell-1} W^{\ell,h}_K}^{T}
    }
    {\sqrt{d/H}}
    +M^{\ell,h}
    },\nonumber
    $}
\end{align}
where $\varphi$ denotes the row-wise softmax function, $\xx^{\ell-1}$ denotes the output of \((\ell-1)\)-th Transformer layer, and $M^{\ell,h}$ is a mask matrix that ensures the attention is causal.
Then, the output of $h$-th multi-head attention can be computed as follows:
\begin{equation}
    \vv^{\ell,h} = A^{\ell,h} \paren{\xx^{\ell-1}W^{\ell,h}_V} \nonumber
\end{equation}
where each $\vv^{\ell,h}_i \in \R^{d}$ is a contextualized value vector of $h$-head at position $i$. 
Once all heads have computed their individual outputs, these outputs are concatenated and passed through the output matrix:
\begin{align}
    \aaa^\ell &= \text{concat}(\vv^{\ell,1}, \cdots, \vv^{\ell,H}) W^{\ell}_O  \nonumber \\
             &= \vv^{\ell} W^{\ell}_O  \nonumber
\end{align}

Compared to the FFN block, multi-head attention directly facilitates information interaction between tokens, and we intervene on the contextualized value vectors.

\section{Method}
Our proposed CP method is a plug-and-play inference-time intervention algorithm that requires no additional data or fine-tuning of the LLM, and can integrate with existing prompt engineering techniques.
CP consists of three steps to obtain sentence embeddings and requires forward propagation for both normal and auxiliary prompts, as illustrated in Figure~\ref{fig:framework}.
In the first step, it inputs the auxiliary prompt into the LLM up to the \(\ell\)-th multi-head attention layer to obtain the non-essential information vector.
Next, it performs forward propagation with the normal prompt and use the non-essential information vector to contrast with it, steering its representation to better emphasize the core semantics.
Finally, we adjust the norm of contextualized value vector before and after the intervention, and forward propagate the adjusted value vectors to extract the sentence embeddings.

\subsection{Auxiliary Prompt}
In the first step, we construct an auxiliary prompt to extract non-essential information in the sentence.
Specifically, we use the auxiliary prompt (i.e., The irrelevant information of this sentence: ``\texttt{[TEXT]}'' means in one word: ``) to wrap the text and feed it into the LLM to obtain the contextualized value vectors from the \(\ell\)-th multi-head attention layer.

Formally, given the input $T_\text{aux} = [t_1, ..., t_{N_\text{aux}}]$ wrapped in the auxiliary template, we pass them into the \(\ell\)-th Transformer layers of LLMs and obtain the auxiliary contextualized value vectors (i.e., $\vv^{\text{aux}, (\ell)} = [\vv^{\text{aux}, (\ell)}_1,\cdots,\vv^{\text{aux}, (\ell)}_{N_{\text{aux}}}]$) from the \(\ell\)-th multi-head attention layer.

The overhead of introducing an auxiliary prompt is minimal, as it only needs to propagate to the \(\ell\)-th layer, rather than all layers.
Additionally, for methods with multiple prompts, compared to each normal prompt requiring propagation through the middle and later layers of the LLM, the auxiliary prompt only needs to propagate through the lower layers of the LLM and can refine all normal prompts with a single auxiliary prompt, further reducing the overhead.

\subsection{Contrastive Activation Steering}
In the second step, it performs forward propagation with the normal prompt and use the non-essential information vectors to intervene and refine the normal prompt.

We also use the normal prompt to wrap the text and feed the input $T_\text{nor} = [t_1, ..., t_{N_\text{nor}}]$ into the LLM to obtain the contextualized value vectors (i.e., $\vv^{\text{nor}, (\ell)} = [\vv^{\text{nor}, (\ell)}_1,\cdots,\vv^{\text{nor}, (\ell)}_{N_{\text{nor}}}]$) from the \(\ell\)-th multi-head attention layer.
The normal prompt can be any existing prompt-based methods, such as PromptEOL~\cite{jiang2023scaling}, Pretended CoT~\cite{zhang2024simple}, and Knowledge~\cite{zhang2024simple}.

Then, we obtain the semantic activation vector $\Delta\vv^{\ell}$ by contrasting the contextualized value vectors derived from the normal and auxiliary prompts.
Intuitively, the result of the difference removes the non-essential information, allowing it to focus more on the core semantics of a sentence.
Specifically, this vector is calculated as:
\begin{align}
    \Delta\vv^{\ell} = \vv^{\text{nor},\paren{\ell}}_{N_{\text{nor}}} - \vv^{\text{aux},\paren{\ell}}_{N_{\text{aux}}}.
    \label{eq:semantic_vector}
\end{align}

The semantic activation vector is sentence-dependent, as the contextualized value vectors (i.e., $\vv^{\text{nor},\paren{\ell}}_{N_{\text{nor}}}$ and $\vv^{\text{aux},\paren{\ell}}_{N_{\text{aux}}}$) for each sentence differ.
Since we focus solely on sentence embeddings and the lengths of the normal and auxiliary prompts may differ, we intervene only on the last token.

\begin{table*}[th]
\centering
\normalsize
\setlength{\tabcolsep}{5pt}
\resizebox{\linewidth}{!}{%
\begin{tabular}{lcccccccccc}
\toprule
\textbf{Method} & \textbf{Params} & \textbf{STS12} & \textbf{STS13} & \textbf{STS14} & \textbf{STS15} & \textbf{STS16} & \textbf{STS-B} & \textbf{SICK-R} & \textbf{Avg.}\\
\midrule
BERT avg & 110M & 30.87 & 59.89 & 47.73 & 60.29 & 63.73 & 47.29 & 58.22 & 52.57\\
BERT prompt$^\dag$ & 110M & 60.96  & 73.83 & 62.18 & 71.54 & 68.68 & 70.60 & 67.16 & 67.85\\
ST5-Enc avg$^\dag$ & 4.8B & 34.97 & 60.19 & 47.59 & 66.40 & 70.62 & 62.83 & 63.57 & 58.02\\
LLaMA2 avg & 7B & 35.49 & 53.15 & 40.12 & 55.35 & 53.26 & 42.10 &      49.96 & 47.06 \\
LLaMA2 echo$^\dag$ & 7B & 52.40 &72.40&61.24&72.67&73.51&65.73& 64.39& 66.05 \\
{MetaEOL}$^\dag$  & 7B& 64.16 & 81.61 & 73.09 & 81.11 & 78.94 & 77.96 & 74.86 & 75.96 \\
\midrule
{PromptEOL}$^\dag$ & 7B & 58.81 & 77.01 & 66.34 & 73.22 & 73.56 & 71.66 & 69.64 & 70.03  \\
{PromptEOL} + CP-NS (\textbf{\textit{Ours}})& 7B &  \cellcolor{tabcolor3} 63.34 \up{4.53} & \cellcolor{tabcolor3} 82.15 \up{5.14}& \cellcolor{tabcolor3} 71.73 \up{5.39}& \cellcolor{tabcolor3} 79.68 \up{6.46}& \cellcolor{tabcolor3} 77.23 \up{3.67}& \cellcolor{tabcolor3} 78.71 \up{7.05}& \cellcolor{tabcolor3} 74.04 \up{4.40} & \cellcolor{tabcolor3} 75.27 \up{5.24} \\
{PromptEOL} + CP-NR (\textbf{\textit{Ours}})& 7B & \cellcolor{tabcolor3} 63.37 \up{4.56} & \cellcolor{tabcolor3} 81.95 \up{4.94}& \cellcolor{tabcolor3} 71.90 \up{5.56}& \cellcolor{tabcolor3} 79.54 \up{6.32}& \cellcolor{tabcolor3} 77.29 \up{3.73}& \cellcolor{tabcolor3} 78.36 \up{6.70}& \cellcolor{tabcolor3} 74.00 \up{4.36} & \cellcolor{tabcolor3} 75.20 \up{5.17} \\
\midrule
Pretended CoT$^\dag$ & 7B& 67.45 & 83.89 & 74.14 & 79.47 & 80.76 & 78.95 & 73.33 & 76.86  \\
Pretended CoT + CP-NS (\textbf{\textit{Ours}})& 7B & \cellcolor{tabcolor3} 67.79 \up{0.34} & \cellcolor{tabcolor3} 83.66 \down{0.23}& \cellcolor{tabcolor3} 74.52 \up{0.38}& \cellcolor{tabcolor3} 81.10 \up{1.63}& \cellcolor{tabcolor3} 80.70 \down{0.06}& \cellcolor{tabcolor3} 80.39 \up{1.44}& \cellcolor{tabcolor3} 74.01 \up{0.68} & \cellcolor{tabcolor3} 77.45 \up{0.59} \\
Pretended CoT + CP-NR (\textbf{\textit{Ours}})& 7B & \cellcolor{tabcolor3} 67.62 \up{0.17} & \cellcolor{tabcolor3} 83.69 \down{0.20}& \cellcolor{tabcolor3} 74.53 \up{0.41}& \cellcolor{tabcolor3} 81.13 \up{1.66}& \cellcolor{tabcolor3} 80.74 \down{0.02}& \cellcolor{tabcolor3} 80.39 \up{1.44}& \cellcolor{tabcolor3} 74.02 \up{0.69} & \cellcolor{tabcolor3} 77.45 \up{0.59} \\
\midrule
Knowledge$^\dag$  & 7B& 65.60 & 82.82 & 74.48 & 80.75 & 80.13 & 80.34 & 75.89 & 77.14\\
Knowledge + CP-NS (\textbf{\textit{Ours}})& 7B & \cellcolor{tabcolor3} 67.16 \up{1.56} & \cellcolor{tabcolor3} 83.43 \up{0.61}& \cellcolor{tabcolor3} 74.23 \down{0.25}& \cellcolor{tabcolor3} 81.29 \up{0.54}& \cellcolor{tabcolor3} 80.03 \down{0.10}& \cellcolor{tabcolor3} 80.80 \up{0.46}& \cellcolor{tabcolor3} 75.97 \up{0.08} & \cellcolor{tabcolor3} 77.56 \up{0.42} \\
Knowledge + CP-NR (\textbf{\textit{Ours}})& 7B & \cellcolor{tabcolor3} 66.65 \up{1.05} & \cellcolor{tabcolor3} 83.21 \up{0.39}& \cellcolor{tabcolor3} 74.21 \down{0.27}& \cellcolor{tabcolor3} 81.19 \up{0.44}& \cellcolor{tabcolor3} 79.74 \down{0.39}& \cellcolor{tabcolor3} 80.70 \up{0.36}& \cellcolor{tabcolor3} 76.07 \up{0.18} & \cellcolor{tabcolor3} 77.40 \up{0.26} \\
\midrule
CK  & 7B& 67.11 & 84.03 & 75.07 & 82.42 & 80.91 & 81.84 & 76.24 & 78.23\\
CK + CP-NS (\textbf{\textit{Ours}})& 7B  & \cellcolor{tabcolor3} 68.35 \up{1.24} & \cellcolor{tabcolor3} 84.21 \up{0.18}& \cellcolor{tabcolor3} 75.59 \up{0.52}& \cellcolor{tabcolor3} 82.49 \up{0.07}& \cellcolor{tabcolor3} 81.50 \up{0.59}& \cellcolor{tabcolor3} 82.29 \up{0.45}& \cellcolor{tabcolor3} 76.34 \up{0.10} & \cellcolor{tabcolor3} 78.68 \up{0.45} \\
CK + CP-NR (\textbf{\textit{Ours}})& 7B  & \cellcolor{tabcolor3} 68.09 \up{0.98} & \cellcolor{tabcolor3} 84.14 \up{0.11}& \cellcolor{tabcolor3} 75.58 \up{0.51}& \cellcolor{tabcolor3} 82.46 \up{0.04}& \cellcolor{tabcolor3} 81.39 \up{0.48}& \cellcolor{tabcolor3} 82.17 \up{0.33}& \cellcolor{tabcolor3} 76.39 \up{0.15} & \cellcolor{tabcolor3} 78.60 \up{0.37} \\
\bottomrule
\end{tabular}
}
\caption{Results on STS tasks using LLaMA2-7B as the backbone. \dag denotes the results from the original paper.}
\label{tab:results}
\end{table*}

\subsection{Norm Adjustment}
In the third step, it adjusts the norm of the contextualized value vector after intervention, and then continues the forward propagation with the adjusted contextualized value vector.
Although we obtain semantic activation vectors, the norm of the contextualized value vector significantly changes after intervention. Thus, we further propose two strategies for adjusting the norm: norm scaling and norm recovering.

\textbf{Norm Scaling (NS)} additionally introduces a hyperparameter to control the norm:
\begin{align}
   \hat{\vv}^{\text{nor},\paren{\ell}}_{N_{\text{nor}}} = \alpha \cdot \Delta\vv^{\ell},
\end{align}
where $\alpha$ is a scaling factor that controls the extent of contrastive activation steering.

\textbf{Norm Recovering (NR)} strategy ensures that the norm remains consistent before and after intervention and avoids the introduction of additional hyperparameters.
NR strategy can ensure that the subsequent output matrix receives inputs similar to the original ones, thereby maintaining the model's capabilities.
Specifically, we renormalize the updated value vector to align with the $L^2\text{-norm}$ before the intervention:
\begin{align}
   \hat{\vv}^{\text{nor},\paren{\ell}}_{N_{\text{nor}}}  = \Delta\vv^{\ell} \cdot \frac{\lVert\vv^{\text{nor},\paren{\ell}}_{N_{\text{nor}}}\rVert_2}{\lVert\Delta\vv^{\ell}\rVert_2}.
\end{align}

After getting $\hat{\vv}^{\text{nor},\paren{\ell}}_{N_{\text{nor}}}$, we further use $\hat{\vv}^{\text{nor},\paren{\ell}}_{N_{\text{nor}}}$ to replace $\vv^{\text{nor},\paren{\ell}}_{N_{\text{nor}}}$ and obtain the new input $\hat{\vv}^{\text{nor}, (\ell)}$ for the output matrix.
Specifically,
\begin{align}
   \hat{\vv}^{\text{nor}, (\ell)} = [\vv^{\text{nor}, (\ell)}_1, \cdots, \hat{\vv}^{\text{nor}, (\ell)}_{N_{\text{nor}}}].
\end{align}

Next, we feed the adjusted contextualized value vector $\hat{\vv}^{\text{nor}, (\ell)}$ into the output matrix $W^\ell_O$ of the \(\ell\)-th multi-head attention layer and continue the standard forward propagation to extract sentence embeddings.

\subsection{Intermediate Embedding Eliciting}
Recent studies~\cite{liu/fantastic,jin/exploring} have confirmed that each layer of the LLM serves a different purpose, with the embeddings extracted from the final layer primarily used for prediction, and they may not yield the best performance.
Thus, we use embeddings extracted from intermediate layers, rather than the final layer, as sentence embeddings.
Extracting embeddings from the intermediate layers not only provides high quality embeddings but also avoids the high cost of propagating through to the final layer, thereby accelerating the extraction of embeddings.
We can use the validation set to determine which layer's embeddings to use, and the overhead of this process is lightweight.

\section{Experiments}

\subsection{Datasets and Experimental Settings}
We evaluate sentence embeddings on seven semantic textual similarity (STS) datasets, including STS 2012-2016~\cite{DBLP:conf/semeval/AgirreCDG12,DBLP:conf/starsem/AgirreCDGG13,DBLP:conf/semeval/AgirreBCCDGGMRW14,DBLP:conf/semeval/AgirreBCCDGGLMM15,DBLP:conf/semeval/AgirreBCDGMRW16}, STS-B~\cite{Cer17}, and SICK-R~\cite{DBLP:conf/lrec/MarelliMBBBZ14}.
Each sentence pair in these datasets is annotated with a pairwise semantic similarity score ranging from 0 to 5.
We use cosine similarity to calculate the predicted similarity scores and evaluate them using Spearman correlation, which assesses the degree of rank correlation between the predicted similarity scores and the annotated similarity scores.

We use grid search on the STS-B development set to search for the intervention layer $\ell$ in \{3, 4, 5, 6, 7\} and the scaling factor of the norm scaling $\alpha$ in \{0.5, 1, 2, 3, 4\} for each prompt.
The intervention layer of PromptEOL is the 5th layer, while the intervention layer of Pretended CoT and Knowledge is the 7th layer.
The norm scaling of PromptEOL is 2, while the norm scaling of Pretended CoT and Knowledge is 3.
We use the 27th layer as the output layer for PromptEOL and Pretended CoT, while the penultimate layer is used for Knowledge.

\begin{table*}[th]
\centering
\normalsize
\setlength{\tabcolsep}{5pt}
\resizebox{\linewidth}{!}{%
\begin{tabular}{lrccccccccc}
\toprule
\textbf{Method} & \textbf{Backbone} & \textbf{STS12} & \textbf{STS13} & \textbf{STS14} & \textbf{STS15} & \textbf{STS16} & \textbf{STS-B} & \textbf{SICK-R} & \textbf{Avg.}\\
\midrule
\midrule
Pretended CoT & LLaMA2-7B & 67.45 & 83.89 & 74.14 & 79.47 & 80.76 & 78.95 & 73.33 & 76.86  \\
Pretended CoT + CP-NS (\textbf{\textit{Ours}})& LLaMA2-7B & \cellcolor{tabcolor3} 66.70 \down{0.75} & \cellcolor{tabcolor3} 84.26 \up{0.37}& \cellcolor{tabcolor3} 74.44 \up{0.30}& \cellcolor{tabcolor3} 81.39 \up{1.92}& \cellcolor{tabcolor3} 80.71 \down{0.05}& \cellcolor{tabcolor3} 80.56 \up{1.61}& \cellcolor{tabcolor3} 74.07 \up{0.74} & \cellcolor{tabcolor3} 77.45 \up{0.59} \\
Pretended CoT + CP-NR (\textbf{\textit{Ours}})& LLaMA2-7B & \cellcolor{tabcolor3} 66.77 \down{0.68} & \cellcolor{tabcolor3} 84.31 \up{0.42}& \cellcolor{tabcolor3} 74.55 \up{0.41}& \cellcolor{tabcolor3} 81.45 \up{1.98}& \cellcolor{tabcolor3} 80.73 \down{0.03}& \cellcolor{tabcolor3} 80.51 \up{1.56}& \cellcolor{tabcolor3} 74.08 \up{0.75} & \cellcolor{tabcolor3} 77.49 \up{0.63} \\
\midrule
Pretended CoT & LLaMA2-13B & 64.27 & 78.61 & 69.93 & 76.37 & 79.28 & 75.88 & 69.04 & 73.34  \\
Pretended CoT + CP-NS (\textbf{\textit{Ours}}) & LLaMA2-13B & \cellcolor{tabcolor3} 64.41 \up{0.14} & \cellcolor{tabcolor3} 78.79 \up{0.18}& \cellcolor{tabcolor3} 69.71 \down{0.22}& \cellcolor{tabcolor3} 76.59 \up{0.22}& \cellcolor{tabcolor3} 79.35 \up{0.07}& \cellcolor{tabcolor3} 77.37 \up{1.49}& \cellcolor{tabcolor3} 71.18 \up{2.14} & \cellcolor{tabcolor3} 73.91 \up{0.57} \\
Pretended CoT + CP-NR (\textbf{\textit{Ours}}) & LLaMA2-13B & 
\cellcolor{tabcolor3} 64.08 \down{0.19} & \cellcolor{tabcolor3} 79.00 \up{0.39}& \cellcolor{tabcolor3} 69.61 \down{0.32}& \cellcolor{tabcolor3} 76.90 \up{0.53}& \cellcolor{tabcolor3} 79.43 \up{0.15}& \cellcolor{tabcolor3} 77.23 \up{1.35}& \cellcolor{tabcolor3} 70.60 \up{1.56} & \cellcolor{tabcolor3} 73.84 \up{0.50} \\
\midrule
Pretended CoT & LLaMA3.1-8B & 61.71 & 81.29 & 69.48 & 77.88 & 78.92 & 76.31 & 72.92 & 74.07  \\
Pretended CoT + CP-NS (\textbf{\textit{Ours}}) & LLaMA3.1-8B & \cellcolor{tabcolor3} 63.25 \up{1.54} & \cellcolor{tabcolor3} 82.55 \up{1.26}& \cellcolor{tabcolor3} 70.73 \up{1.25}& \cellcolor{tabcolor3} 79.16 \up{1.28}& \cellcolor{tabcolor3} 80.06 \up{1.14}& \cellcolor{tabcolor3} 77.52 \up{1.21}& \cellcolor{tabcolor3} 73.28 \up{0.36} & \cellcolor{tabcolor3} 75.22 \up{1.15} \\
Pretended CoT + CP-NR (\textbf{\textit{Ours}}) & LLaMA3.1-8B & \cellcolor{tabcolor3} 63.74 \up{2.03} & \cellcolor{tabcolor3} 82.46 \up{1.17}& \cellcolor{tabcolor3} 70.59 \up{1.11}& \cellcolor{tabcolor3} 78.92 \up{1.04}& \cellcolor{tabcolor3} 80.10 \up{1.18}& \cellcolor{tabcolor3} 77.45 \up{1.14}& \cellcolor{tabcolor3} 73.35 \up{0.43} & \cellcolor{tabcolor3} 75.23 \up{1.16} \\
\bottomrule
\end{tabular}
}
\caption{Results on STS tasks (Spearman correlation scaled by 100x) using different backbones. Since Pretended CoT generalizes better across different LLMs, we use it for the experiment.}
\label{tab:backbone}
\end{table*}

\subsection{Baselines}
We combine our method with some baselines to demonstrate effectiveness.
\textbf{BERT avg}~\cite{DBLP:conf/naacl/DevlinCLT19}, \textbf{ST5-Enc avg}~\cite{ni2022sentence}, and \textbf{LLaMA2 avg}~\cite{touvron2023llama} generate sentence embeddings by averaging all token embeddings, each utilizing a different backbone.
\textbf{BERT prompt}~\cite{DBLP:conf/emnlp/JiangJHZWZWHDZ22} proposes to represent a sentence with a prompt using BERT.
\textbf{LLaMA2 echo}~\cite{springer2024repetition} repeats the input twice in context and uses mean-pooling to extract embeddings from the second occurrence.
\textbf{PromptEOL}~\cite{jiang2023scaling} compresses the semantics of a sentence into a single token to extract embeddings.
\textbf{MetaEOL}~\cite{lei2024meta} leverages a diverse set of meta-task prompts to capture multiple representations of sentences from distinct perspectives.
\textbf{Pretended CoT}~\cite{zhang2024simple} uses CoT to inspire LLMs to extract better sentence embeddings.
\textbf{Knowledge}~\cite{zhang2024simple} provides explicit guidance to LLMs by conveying human experience in text summarization through prompts.
We additionally create a multi-prompt baseline \textbf{CK}, using the average of the embeddings extracted by \textbf{CoT} and \textbf{Knowledge} as the sentence embedding.
The detailed prompts for \textbf{PromptEOL}, \textbf{CoT}, and \textbf{Knowledge} are shown in Appendix A.

\begin{table}[t] \small
    \centering
    \setlength{\tabcolsep}{13pt}
    \begin{tabular}{lcc}
        \toprule
        \textbf{Method} & \textbf{w/o CP} & \textbf{w/ CP}\\
        \midrule
        \textbf{PromptEOL}  & 27 ($1\times$)  & 31 ($1.15\times$)\\
        \textbf{Pretended CoT} & 27 ($1\times$) &  31 ($1.15\times$)\\
        \textbf{Knowledge} & 31 ($1.15\times$) & 37 ($1.37\times$) \\
        \textbf{CK} & 54 ($2\times$) & 60 ($2.22\times$) \\
        \bottomrule
    \end{tabular}
    \caption{The number of layers in forward propagation.}
    \label{tab:time}
\end{table}

\begin{table*}[th]
    \centering
    \small
    \setlength{\tabcolsep}{25pt}
    \begin{tabular}{l c}
        \toprule
        \textbf{Prompt} & \textbf{Knowledge+CP-NS} \\
        \midrule
        The irrelevant information of this sentence: ``\texttt{[TEXT]}'' means in one word:\" &  82.26 \up{0.43} \\
        The redundant information of this sentence: ``\texttt{[TEXT]}'' means in one word:\" &  82.41 \up{0.58} \\
        The background of this sentence: ``\texttt{[TEXT]}'' means in one word:\" &  82.12 \up{0.29} \\ 
        The descriptive term of this sentence: ``\texttt{[TEXT]}'' means in one word:\" & 82.48 \up{0.65} \\
        \midrule
        The sentence: ``\texttt{[TEXT]}'' reflects the sentiment in one word:\" & 77.17 \down{4.66} \\
        The sentence: ``\texttt{[TEXT]}'' highlights the primary entity or relation in one word:\" & 81.62 \down{0.21} \\
        \bottomrule
    \end{tabular}
    \caption{Effects of auxiliary prompt on the STS-B development set.}
    \label{tab:auxiliary}
\end{table*}

\begin{table}[t] \small
    \centering
    \setlength{\tabcolsep}{18pt}
    \begin{tabular}{lcccc}
        \toprule
        \textbf{Position} & \textbf{\textbf{Knowledge+CP-NS}}\\
        \midrule
        \textbf{Head} &  82.61 \up{0.78}\\
        \textbf{FFN} & 81.93 \up{0.10}\\
        \textbf{Hidden} & 82.53 \up{0.70}\\
        \bottomrule
    \end{tabular}
    \caption{Effects of intervention position on the STS-B development set. FFN denotes the output of FFN block, and Hidden denotes the output of the Transformer layer.}
    \label{tab:position}
\end{table}

\subsection{Results}
The results of our method on the STS benchmark are shown in Table \ref{tab:results}.
Our method shows improvement in 48 out of 56 cases, surpassing previous methods in average performance.
This indicates that our method can effectively steer the existing prompt-based methods to focus more on semantics.
Compared to the norm recovering strategy, the norm scaling strategy typically achieves better performance.
This suggests that maintaining the consistency of the norm before and after the intervention is not essential for extracting sentence embeddings.

Our method achieves the greatest improvement on PromptEOL.
This is because PromptEOL is the simplest compared to the other prompts, which limits its semantic encoding ability and makes it more necessary to refine it with auxiliary prompts.
Our method can narrow the gap between different prompts, to some extent avoiding the variance caused by different prompts.

Finally, our method can improve the performance of CK, which demonstrates that our approach is still effective for averaging the embeddings of multiple prompts.
This indicates that averaging the embeddings of multiple prompts still contains non-essential information.
In addition, our method combined with CK achieves improvements on all datasets.
This indicates that combining multiple prompts makes the performance improvement more robust.

\begin{figure*}[t]
    \centering
    \includegraphics[width=1\textwidth]{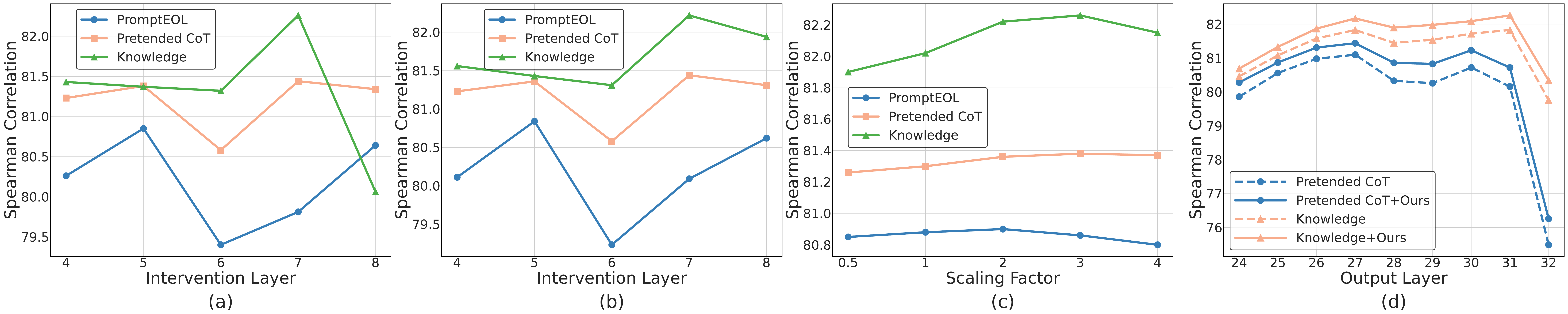}
    \caption{Effects of intervention layer, scaling factor, and output layer. (a) Effects of the intervention layer on norm scaling. (b) Effects of the intervention layer on norm recovering. (c) Effects of the scaling factor on norm scaling. (d) The effects of the output layer on Knowledge+CP-NS.}
    \label{fig:layer_analysis}
\end{figure*}

\begin{table*}[t]
\centering
\normalsize
\setlength{\tabcolsep}{5pt}
\resizebox{\linewidth}{!}{%
\begin{tabular}{lcccccccccc}
\toprule
\textbf{Method} & \textbf{Params} & \textbf{MR} & \textbf{CR} & \textbf{SUBJ} & \textbf{MPQA} & \textbf{SST2} & \textbf{TREC} & \textbf{MRPC} & \textbf{Avg.}\\
\midrule
\midrule
\multicolumn{10}{l}{\it{Fine-tuning on supervised datasets}}\\
SimCSE-RoBERTa & 123M & 84.92 &92.00 &94.11 &89.82 &91.27 & 88.80 &75.65 & 88.08  \\
ST5-Enc & 4.8B & 90.83 &94.44 & 96.33 & 91.68 & 94.84 & 95.40 &77.91 & 91.63  \\
\midrule
\midrule
\multicolumn{10}{l}{\it{Without fine-tuning}}\\
PromptEOL & 7B & 90.63 &92.87 &96.32 &91.19 &95.00 &95.40 &75.19 &90.94  \\
PromptEOL + CP-NS (\textbf{\textit{Ours}}) & 7B &  \cellcolor{tabcolor3} 90.49 \down{0.14} & \cellcolor{tabcolor3} 93.11 \up{0.24}& \cellcolor{tabcolor3} 96.97 \up{0.65}& \cellcolor{tabcolor3} 91.10 \down{0.09}& \cellcolor{tabcolor3} 95.94 \up{0.94}& \cellcolor{tabcolor3} 97.00 \up{1.60}& \cellcolor{tabcolor3} 77.51 \up{2.32} & \cellcolor{tabcolor3} 91.73 \up{0.79} \\
PromptEOL + CP-NR (\textbf{\textit{Ours}}) & 7B &  \cellcolor{tabcolor3} 90.33 \down{0.30} & \cellcolor{tabcolor3} 92.64 \down{0.23}& \cellcolor{tabcolor3} 96.74 \up{0.42}& \cellcolor{tabcolor3} 91.11 \down{0.08}& \cellcolor{tabcolor3} 95.83 \up{0.83}& \cellcolor{tabcolor3} 96.40 \up{1.00}& \cellcolor{tabcolor3} 76.00 \up{0.81} & \cellcolor{tabcolor3} 91.29 \up{0.35} \\
\bottomrule
\end{tabular}
}
\caption{Accuracy on transfer learning tasks using LLaMA2-7B.}
\label{tab:transfer_tasks}
\end{table*}

\begin{table*}[t]
\centering
\small
\setlength{\tabcolsep}{8pt} 
\renewcommand{\arraystretch}{1.5} 
\resizebox{\linewidth}{!}
{%
\begin{tabular}{l l l} 
\toprule
\textbf{Sentence} & \textbf{Method} & \textbf{Top-predicted Tokens and Probability} \\
\midrule
\multirow{4}{*}{\begin{tabular}{@{}c@{}}It is the first named storm to\\ develop in the Caribbean in December.\end{tabular}}
& \multirow{2}{*}{Knowledge}  & It (0.0747), it (0.0233), ir (0.0231), I (0.0221) \\
& & Dec (0.0219), It (0.0211), The (0.0192), What (0.0140) \\
\cline{2-3}
& \multirow{2}{*}{Knowledge+CP-NS} & Dec (0.1092), St (0.0566), H (0.0537), It (0.0460) \\
& & First (0.0391), Car (0.0360), first (0.0289), The (0.0289) \\
\midrule
\multirow{4}{*}{\begin{tabular}{@{}c@{}}The rule - When in doubt throw it out!\end{tabular}}
& \multirow{2}{*}{Knowledge}  & Don (0.0401), Throw (0.0355), Do (0.0337), When (0.0303) \\
& & The (0.0297), throw (0.0206), D (0.0181), If (0.0140) \\
\cline{2-3}
& \multirow{2}{*}{Knowledge+CP-NS} & Throw (0.2071), throw (0.1414), Th (0.0466), D (0.0403) \\
& & Th (0.0341), th (0.0226), TH (0.0185), Don (0.0177) \\
\bottomrule
\end{tabular}}
\caption{Top-8 tokens predicted by different methods using LLaMA2-7B.}\label{tab:case}
\end{table*}

\subsection{Effects of Different Backbones}
Table \ref{tab:backbone} shows the performance across various model backbones, including LLaMA2-7B~\cite{touvron2023llama}, LLaMA2-13B~\cite{touvron2023llama}, and LLaMA3.1-8B \cite{dubey2024llama}.

The results demonstrate that our method can achieve performance improvements across various LLMs, highlighting its generalizability.
Additionally, similar to the previous findings~\cite{lei2024meta}, LLaMA2-13B and LLaMA3.1-8B do not achieve better performance than LLaMA2-7B.
This may be because different LLMs require different prompts to achieve optimal performance.


\subsection{Analysis of the Number of Forward Propagation Layers}
We further report the number of layers of forward propagation to estimate the time overhead in Table~\ref{tab:time}.
The overhead of introducing auxiliary prompts is minimal, as they only need to propagate to the lower layers of the LLM, such as the 5th or 7th layer.
The multi-prompt method CK often has a significant time overhead because each normal prompt needs to propagate to the higher layers of the LLM, whereas our auxiliary prompt only needs to propagate once to affect all the normal prompts, which further reduces the time overhead of our method.

\subsection{Model Analysis}
We further analyze the effects of auxiliary prompt, intervention position, scaling factor, and output layers.

\textbf{Effects of Auxiliary Prompt}
We further explore the effects of auxiliary prompts using Knowledge + NP-NS on the STS-B development set in Table~\ref{tab:auxiliary}.

Using the first five semantically relevant prompts as auxiliary prompts often improves performance.
This suggests that our method is not sensitive to auxiliary prompts.
However, when we use other prompts that focus on sentiment and entities, they often do not improve performance.
This is because these prompts do not focus on non-essential information, but instead focus on certain attributes of the sentence.

\textbf{Effects of the Intervention Position}
We further explore the effects of intervention position and layer.
Intervening in all three positions improves performance, indicating the effectiveness of the intervention, as shown in Table~\ref{tab:position}.
Among them, the intervention on the multi-head attention achieves the best performance.
This may be because the information interaction between tokens primarily occurs in the multi-head attention layer~\cite{elhage2021mathematical}.

The optimal intervention layer for PromptEOL is the 5th layer, while Pretended CoT and Knowledge are the 7th layer, as shown in Figure~\ref{fig:layer_analysis}(a) and (b).
In addition, the optimal intervention layer for both NS and NR is consistent under each prompt.

\textbf{Effects of Scaling Factor}
We investigate the effects of the scaling factor in LLaMA2-7B using three prompts, as shown in Figure \ref{fig:layer_analysis}(c).
The optimal scaling factor for PromptEOL is 2, while Pretended CoT and Knowledge are 3.
As the scaling factor increases, all three prompts show an initial increase followed by a decrease.
When the factor is 2 or 3, NS can achieve good performance across all three prompts.

\textbf{Effects of Output Layers}
We investigate the effects of output layers in LLaMA2-7B using Pretended CoT and Knowledge prompt, as shown in Figure \ref{fig:layer_analysis}(d).
CP-NS consistently improves both Pretended CoT and Knowledge across all layers, with more significant gains in the deeper layers.
This suggests that our method can enhance the embeddings of all middle and later layers in the LLM, rather than just a single layer.
Similar to previous findings~\cite{li2024bellm, lei2024meta}, the sentence embedding of the last layer is not optimal for the STS tasks.
Pretended CoT achieves optimal performance at the 27th layer, and Knowledge is at the 31st layer.
This variation indicates that the optimal output layer varies for different prompts.

\subsection{Transfer Learning Tasks}
We further evaluate the performance of our method on transfer learning tasks, utilizing the standard transfer learning tasks provided by SentEval, in line with prior works~\cite{gao2021simcse,lei2024meta}.
These tasks include MR~\cite{pang2005seeing}, CR~\cite{hu2004mining}, SUBJ \cite{pang2004sentimental}, MPQA~\cite{wiebe2005annotating}, SST-2~\cite{socher2013recursive}, TREC~\cite{voorhees2000building}, and~MRPC \cite{dolan2005automatically}.
For each task, we use the sentence embeddings generated by our method as features to train logistic regression classifiers.
We additionally include two supervised contrastive trained models (SimCSE and ST5-Enc) for reference.
Notably, ST5-Enc, with 4.8 billion parameters, is extensively trained on natural language inference (NLI) data and two billion question-answer pairs.

The results of our method on the transfer learning tasks are shown in Table~\ref{tab:transfer_tasks}.
Our method both outperforms previous methods in average performance, and the norm scaling strategy outperforms the supervised method ST5-Enc.
This further highlights the superiority of our method, which can outperform a 4.8B model trained with supervised contrastive learning, without requiring any additional training.
On relatively simple sentiment classification datasets (such as MR, CR, MPQA, and SST2), the improvement is relatively limited.
However, on more challenging tasks (such as SUBJ, TREC, and MRPC), the improvement is more significant.
This is because different tasks require different embeddings.
Complex tasks require embeddings with stronger semantic understanding capabilities, while for simpler sentiment classification, embeddings extracted from PromptEOL can perform well.

Additionally, the intervention layer is the 7th layer, and the length of norm scaling is 4.
This illustrates the generalizability of the intervention layer, which can avoid extensive hyperparameter search.

\subsection{Case Study}
We further show the top-8 tokens predicted by different methods in Table~\ref{tab:case}.
The example illustrates that Knowledge creates sentence embeddings focusing on stop-word tokens (such as it, I, the, what, Do, When), which convey non-essential information.
In contrast, our method decodes the essential tokens of the sentence, such as Dec, St, First, Car, and Throw.
This further intuitively demonstrates the effectiveness of our method.

\section{Conclusion}
In this paper, we introduce Contrastive Prompting, a plug-and-play inference-time steering method, to extract high-quality sentence embeddings from LLMs without the need for training or additional data.
CP additionally introduces an auxiliary prompt to contrast with the normal prompt, steering it to encode the core semantics of the sentence, rather than non-essential information.
Extensive experiments show that our method can effectively elicit sentence embeddings across a range of diverse LLMs with varying sizes on both STS and transfer learning tasks.

\section*{Limitations}
Firstly, our method preliminarily attempts to use auxiliary prompts to refine the representation of normal prompts.
It is worth further exploring how to generate the optimal auxiliary prompt for each normal prompt.
Secondly, the NR strategy needs to search for a hyperparameter (i.e., intervention layer), and the NS strategy needs to search for two hyperparameters (i.e., intervention layer and scaling factor) to elicit better sentence embeddings. 
Fortunately, in most cases, intervening at the 5th or 7th layer can achieve good performance.
Finally, we conduct experiments only on the English dataset, and we plan to explore this in more languages in the future.

\section*{Acknowledgments}
We would like to thank the anonymous reviewers for their insightful comments.
This work is supported by the National Natural Science Foundation of China under Grants Nos. 62441225, 61972192, 62172208, 61906085.
This work is partially supported by Collaborative Innovation Center of Novel Software Technology and Industrialization.
This work is supported by the Fundamental Research Funds for the Central Universities under Grant No. 14380001.

\bibliography{ref}

\clearpage
\newpage
\appendix

\begin{table*}[!htbp]
\centering
\normalsize
\setlength{\tabcolsep}{5pt}
\resizebox{\linewidth}{!}{%
\begin{tabular}{lcccccccccc}
\toprule
\textbf{Method}  & \textbf{STS12} & \textbf{STS13} & \textbf{STS14} & \textbf{STS15} & \textbf{STS16} & \textbf{STS-B} & \textbf{SICK-R} & \textbf{Avg.}\\
\midrule
{PromptEOL+ICL}$^\dag$ \cite{jiang2023scaling} & 70.65 & 84.51 & 75.01 & 83.51 & 82.00 & 81.12 & 76.77 & 79.08  \\
{PromptEOL+ICL} + CP-NS (\textbf{\textit{Ours}}) &  \cellcolor{tabcolor3} 71.44 \up{0.79} & \cellcolor{tabcolor3} 84.82 \up{0.31}& \cellcolor{tabcolor3} 75.42 \up{0.41}& \cellcolor{tabcolor3} 84.29 \up{0.78}& \cellcolor{tabcolor3} 82.51 \up{0.51}& \cellcolor{tabcolor3} 81.85 \up{0.73}& \cellcolor{tabcolor3} 75.30 \down{1.47} & \cellcolor{tabcolor3} 79.38 \up{0.30} \\
{PromptEOL+ICL} + CP-NR (\textbf{\textit{Ours}}) & \cellcolor{tabcolor3} 70.61 \down{0.04} & \cellcolor{tabcolor3} 84.61 \up{0.10}& \cellcolor{tabcolor3} 75.24 \up{0.23}& \cellcolor{tabcolor3} 83.69 \up{0.18}& \cellcolor{tabcolor3} 82.16 \up{0.16}& \cellcolor{tabcolor3} 81.41 \up{0.29}& \cellcolor{tabcolor3} 76.72 \down{0.05} & \cellcolor{tabcolor3} 79.21 \up{0.13} \\
\bottomrule
\end{tabular}
}
\caption{Results on STS tasks using OPT-6.7B as the backbone. \dag denotes the results from the original paper.}
\label{tab:few-shot}
\end{table*}

\section{Baselines}
We report the detailed prompts for the baselines as follows:
\begin{tcolorbox}
\textbf{PromptEOL:}  
This sentence: ``\texttt{[TEXT]}'' means in one word: ``

\textbf{Pretended CoT:}
After thinking step by step, this sentence: ``\texttt{[TEXT]}'' means in one word: ``

\textbf{Knowledge:}
 The essence of a sentence is often captured by its main subjects and actions, while descriptive terms provide additional but less central details. With this in mind , this sentence: ``\texttt{[TEXT]}'' means  in one word: ``

\end{tcolorbox}

\section{Performance on STS Tasks under In-Context Learning}
In this section, we further explore whether CP can enhance sentence embedding under the in-context learning setting.

We observe that CP also improves performance under in-context learning settings in Table~\ref{tab:few-shot}, demonstrating its generalizability.
The gains are less pronounced compared to the zero-shot scenario, possibly because the additional context already guides the model to focus on the core semantics of the sentence.

\section{Effects of Hyperparameters across Domains}
In this section, we further explore the effect of hyperparameters on the results across different domains, including the STS-12, STS-13, and STS-14 datasets. 

The optimal intervention layer for Knowledge is the 7th layer on all three datasets in Table~\ref{tab:domain}. The optimal scaling factors for the three datasets are 4, 4, and 2, respectively. Therefore, these two hyperparameters do not vary significantly across different domains.

\begin{table}[t] \small
    \centering
    \setlength{\tabcolsep}{8pt}
    \begin{tabular}{lccccc}
        \toprule
        \textbf{STS12} & \textbf{Layer = 6} & \textbf{ Layer = 7} & \textbf{ Layer = 8}\\
        \midrule
        \textbf{$\alpha$ = 1}  & 65.48 & 66.13 & 66.39 \\
        \textbf{$\alpha$ = 2} & 65.44 & 66.74 & 66.61 \\
        \textbf{$\alpha$ = 3} & 65.57 & 67.16 & 66.78 \\
        \textbf{$\alpha$ = 4} & 65.57 & \textbf{67.23} & 66.85 \\
        \textbf{$\alpha$ = 5} & 65.64 & 66.93 & 66.79 \\
        \midrule
        \textbf{STS13} & \textbf{Layer = 6} & \textbf{ Layer = 7} & \textbf{ Layer = 8}\\
        \midrule
        \textbf{$\alpha$ = 1}  & 83.37 & 82.95 & 82.97 \\
        \textbf{$\alpha$ = 2} & 83.33 & 83.24 & 83.01 \\
        \textbf{$\alpha$ = 3} & 83.22 & 84.43 & 83.06 \\
        \textbf{$\alpha$ = 4} & 83.08 & \textbf{84.52} & 83.14 \\
        \textbf{$\alpha$ = 5} & 82.96 & 83.38 & 83.17 \\
        \midrule
        \textbf{STS14} & \textbf{Layer = 6} & \textbf{ Layer = 7} & \textbf{ Layer = 8}\\
        \midrule
        \textbf{$\alpha$ = 1}  & 73.72 & 73.92 & 73.99 \\
        \textbf{$\alpha$ = 2} & 73.61 & \textbf{74.24} & 73.92 \\
        \textbf{$\alpha$ = 3} & 73.52 & 74.23 & 73.84 \\
        \textbf{$\alpha$ = 4} & 73.44 & 74.09 & 74.81 \\
        \textbf{$\alpha$ = 5} & 73.37 & 73.82 & 73.82 \\
        \midrule
        \bottomrule
    \end{tabular}
    \caption{Effects of hyperparameters across domains. Layer denotes the intervention layer.}
    \label{tab:domain}
\end{table}

\section{Multi-Task Evaluation}
We further evaluate the CP method across classification task, clustering task, reranking task, and pair classification task from the MTEB benchmark \cite{muennighoff2022mteb}.
Due to the large size of the MTEB dataset, we evaluate the effectiveness of our method on only a subset of the data.

\begin{table}[th]
    \centering
    \begin{center}
    \setlength{\tabcolsep}{1pt}
    \begin{small}
    \begin{tabular}{c|ccc|cccccccc}
    \toprule
    \textbf{Method} & \textbf{PromptEOL} & \textbf{PromptEOL+CP-NS}      \\
    \midrule
    AmazonCounterfactual      & 70.83     &  \cellcolor{tabcolor3} 73.75 \up{2.92}\\
    Banking77                 & 78.94  & \cellcolor{tabcolor3} 81.54 \up{2.60}\\
    Emotion                   & 48.35   & \cellcolor{tabcolor3} 50.96 \up{2.61}\\
     
    \midrule
    \bf{Average (3)}         & 66.04   & \cellcolor{tabcolor3} 68.75 \up{2.71}\\
    \bottomrule
    \end{tabular}
    \caption{Accuracy on classification datasets using LLaMA2-7B.}
    \label{table:mteb_classification}
    \end{small}
    \end{center}
\end{table}

\begin{table}[t] \scriptsize
    \begin{center}
    \setlength{\tabcolsep}{7pt}
    \begin{tabular}{c|ccc|cccccccc}
    \toprule
    \textbf{Method} & \textbf{PromptEOL} & \textbf{PromptEOL+CP-NS}      \\
    \midrule
    SprintDuplicateQuestions  & 43.02    & \cellcolor{tabcolor3}48.60 \up{5.58}\\
    TwitterSemEval2015        & 65.61   & \cellcolor{tabcolor3}68.55 \up{2.94}\\
    \midrule
    \bf{Average (2)}         & 54.32  & \cellcolor{tabcolor3} 58.58 \up{4.26}\\
    \bottomrule
    \end{tabular}
    \caption{Accuracy on pair classification datasets using LLaMA2-7B.}
    \label{table:mteb_pair_classification}
    \end{center}
\end{table}

\begin{table}[th] 
    \begin{center}
    \setlength{\tabcolsep}{5pt}
    \scriptsize
    \begin{tabular}{c|ccc|cccccccc}
    \toprule
    \textbf{Method} & \textbf{PromptEOL} & \textbf{PromptEOL+CP-NS}      \\
    \midrule
    AskUbuntuDupQuestions     & 53.88 & \cellcolor{tabcolor3} 57.02 \up{3.14} \\
    SciDocsRR                 & 71.38 & \cellcolor{tabcolor3} 77.94 \up{6.56}\\
    StackOverflowDupQuestions & 40.63  & \cellcolor{tabcolor3} 43.04 \up{2.41}\\ 
    \midrule
    \bf{Average (3)}         & 55.30   & \cellcolor{tabcolor3} 59.33 \up{4.03}\\
    \bottomrule
    \end{tabular}
    \caption{Average precision on reranking datasets using LLaMA2-7B.}
    \label{table:mteb_reranking}
    \end{center}
\end{table}

Our method consistently improves performance across classification, pair classification, reranking, and clustering tasks, as demonstrated in Tables \ref{table:mteb_classification}, \ref{table:mteb_pair_classification}, \ref{table:mteb_reranking}, and \ref{table:mteb_clustering}.

\begin{table}[t] \scriptsize
    \begin{center}
    \setlength{\tabcolsep}{5pt}
    \begin{tabular}{c|ccc|cccccccc}
    \toprule
    \textbf{Method} & \textbf{PromptEOL} & \textbf{PromptEOL+CP-NS}      \\
    \midrule
    
    TwentyNewsgroupsClustering &27.61                        &\cellcolor{tabcolor3} 35.53 \up{7.92} \\
    \bottomrule
    \end{tabular}
    \caption{V-measure on clustering datasets using LLaMA2-7B.}
    \label{table:mteb_clustering}
    \end{center}
\end{table}

\end{document}